%% file: main.tex
\documentclass[10pt,twocolumn,letterpaper]{article}

\usepackage{algorithmic}
\usepackage{amsmath}

\usepackage{cvpr}              
\usepackage{multirow}
\usepackage{adjustbox}
\usepackage{bbm}
\usepackage{wrapfig,lipsum}
\usepackage{xspace}
\usepackage{algorithm}
\usepackage{amsfonts}
\usepackage[accsupp]{axessibility}  

\definecolor{cvprblue}{rgb}{0.21,0.49,0.74}

\usepackage[pagebackref,breaklinks,colorlinks,allcolors=cvprblue]{hyperref}

\def\paperID{***} 
\def\confName{CVPR}
\def\confYear{2026}

\title{DRS-GUI: Dynamic Region Search for Training-Free GUI Grounding}

\author{
Yichao Liu\textsuperscript{1} \quad
Huawen Shen\textsuperscript{2} \quad
Liu Yu\textsuperscript{1} \quad
Shiyu Liu\textsuperscript{1} \quad
Zeyu Chen\textsuperscript{1} \quad
Yu Zhou\textsuperscript{1}\thanks{Corresponding author.} \\
\textsuperscript{1}Nankai University\\
\textsuperscript{2}Institute of Information Engineering, Chinese Academy of Sciences\\
}

\begin{document}
\maketitle
\input{sec/0_abstract}    
\input{sec/1_intro}
\input{sec/2_related_works}

\input{sec/3_method}

\input{sec/4_experiments}
\input{sec/5_conclusion}
{
    \small
    \bibliographystyle{ieeenat_fullname}
    \bibliography{main}
}


\end{document}

%% file: sec/0_abstract.tex
\begin{abstract}
GUI agents powered by Multimodal Large Language Models (MLLMs) have demonstrated impressive capability in understanding and executing user instructions. However, accurately grounding instruction-relevant elements from high-resolution screenshots cluttered with irrelevant UI components remains challenging for existing approaches. Inspired by how humans dynamically adjust their perceptual scope to locate task-related regions on complex screens, we propose DRS-GUI, a training-free dynamic region search framework for GUI grounding that can be seamlessly integrated into existing MLLMs. DRS-GUI introduces a lightweight UI Perceptor that performs three human-like perceptual actions (Focus, Shift, and Scatter) to progressively explore the interface and generate region proposals. To dynamically schedule these actions, we further design an Action Planner based on Monte Carlo Tree Search (MCTS). A region quality reward is employed to evaluate and select the highly instruction-relevant region, efficiently pruning redundant UI elements. Experiments demonstrate that DRS-GUI yields a 14\% improvement on ScreenSpot-Pro for general and GUI-specific MLLMs (Qwen2.5-VL-7B and UGround-V1-7B), significantly enhancing grounding  performance and generalization.




\end{abstract}

%% file: sec/1_intro.tex
\section{Introduction}
\label{sec:intro}





\begin{figure}[t]
    \centering
    \includegraphics[width=\linewidth]{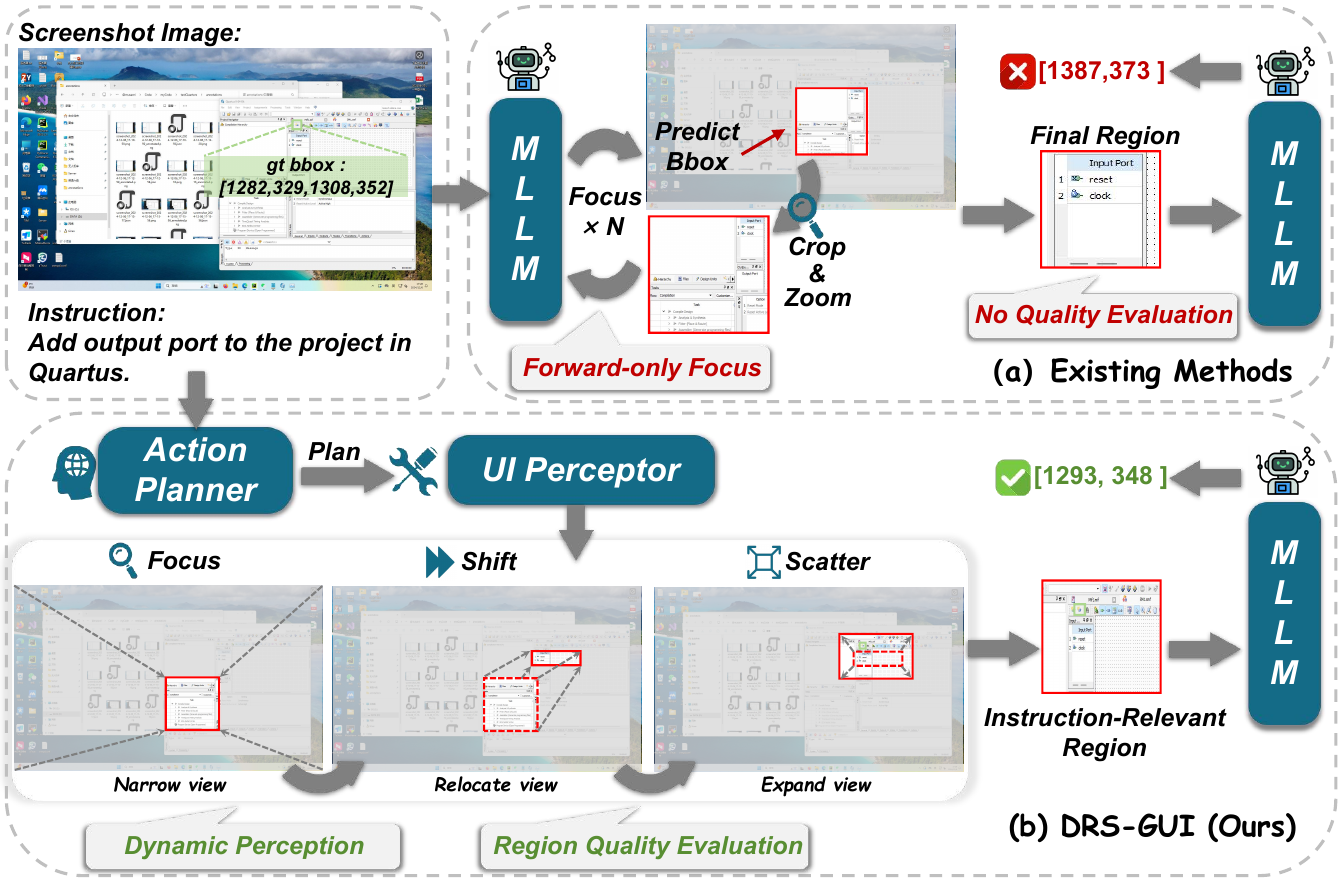}
    \caption{\textbf{(a)} Single-stage methods progressively crop and zoom through forward-only focus, causing error accumulation. \textbf{(b)} DRS-GUI uses an action planner and a UI Perceptor to explore regions via three perceptual actions, guided by region quality evaluation to select the instruction-relevant region.}
    \label{fig:Overview}
    \vspace{-12pt}
\end{figure}

Graphical user interface (GUI) agents, powered by recent advances in multimodal large language models (MLLMs) ~\cite{xu2024aguvis, gou2024navigating, yang2025aria, wu2024atlas, shen2024falcon}, are transitioning from rule-based scripted control to natural language interaction. The reliability of such interaction fundamentally depends on \emph{grounding}-accurately localizing the executable UI element specified by a user instruction \cite{lin2024showui, yuan2025enhancing}. However, real GUI screenshots are often high-resolution and visually dense, with substantial redundant elements that distract MLLMs from the instruction-relevant cues needed to reliably isolate the true target~\cite{li2025screenspot}. Resulting failures arise not only from visual clutter, but also from the absence of an explicit mechanism for adapting perceptual scope during interaction.

Existing GUI grounding approaches typically fall into two paradigms, yet both lack truly adaptive perceptual behavior. The first paradigm formulates grounding as a single-step full-screen prediction, directly regressing a point or bounding box over the entire screenshot~\cite{leigrounding, seeclick, wu2025gui}. Without explicit control over attention, such models often fail to isolate instruction-relevant cues from background clutter. The second paradigm adopts multi-step refinement strategies, such as iterative cropping or zooming~\cite{lee2025training, park2025r, wang2025learning}, to gradually narrow the view toward the target. Although these methods provide progressive focus, their refinement processes are typically forward-only and lack mechanisms to assess whether each step improves alignment with the instruction or to backtrack when early decisions deviate from the correct trajectory~\cite{wu2025dimo}. As illustrated in~\cref{fig:Overview} (a), when the trajectory drifts away from semantic cues, the final region may fail to contain the target, making the error irrecoverable.

In contrast, human behavior follows a more flexible pattern: rather than remaining on a single narrowing path, 
humans scan the layout, refine attention gradually, and step back or shift attention to alternative regions when uncertainty arises. 
This adaptive and reversible search process naturally evaluates whether attention is moving toward or away from the true target~\cite{torralba2006contextual, wolfe2020visual, wolfe2011visual}, a capability that current GUI grounding pipelines lack.

These observations suggest that a reliable GUI grounding framework should exhibit two key capabilities:
(1) \textbf{dynamic perception}, the ability to revise or shift the viewing region when current evidence is weak rather than committing to a single irreversible narrowing path; and 
(2) \textbf{region quality evaluation}, where each perceptual step provides a measurable signal indicating whether the search is moving closer to the target, helping to prevent early errors from compounding. To realize these capabilities, we propose \textbf{DRS-GUI}, a training-free framework that introduces a dynamic region search and decision procedure to guide the grounding process. 

DRS-GUI dynamically adjusts the perceptual scope by proposing candidate regions, evaluating their relevance to the instruction through semantic and structural cues, and selecting refinement or redirection actions when necessary. This adaptive perceptual process echoes how humans selectively explore and adjust attention within visually dense interfaces. To support this process, DRS-GUI incorporates an auxiliary \emph{UI Perceptor}, built upon a UI element parser \cite{lu2024omniparser} and a text-embedding module \cite{su2022one}. The Perceptor provides structured UI representations that help determine which regions carry instruction-relevant cues. Guided by these cues, the system executes three perceptual actions: Focus (contract the view for finer detail), Shift (relocate attention to alternative regions), and Scatter (expand outward to recover broader context).
To dynamically schedule these actions, DRS-GUI employs an \emph{Action Planner} that integrates Monte Carlo Tree Search (MCTS)~\cite{silver2016mastering} to enable human-like planning of field-of-view adjustments.
We further design a reward that evaluates region quality and determines whether a candidate region should be further explored or selected as final. This step-wise evaluation mitigates the accumulation of early errors, addressing the inherent brittleness of purely forward refinement. DRS-GUI requires no additional model training or fine-tuning, making it a plug-and-play enhancement for existing MLLMs~\cite{bai2025qwen2, chen2023internvl}. By dynamically generating and evaluating perceptual regions, DRS-GUI reduces redundant visual clutter and improves grounding robustness on high-resolution, element-dense interfaces. In summary, our contributions are as follows:
\begin{itemize}
    \item We propose DRS-GUI, a training-free GUI grounding approach that performs dynamic region search, effectively enhancing MLLMs grounding performance.
    \item We introduce a human-like perception–action process to obtain a reliable target region: UI Perceptor provides instruction-relevant cues to drive three perceptual actions, and an MCTS-based Action Planner with a region quality reward regulates these actions to reach a stable step-wise decision.
    \item 
    Experiments show that DRS-GUI achieves significant improvements on the high-resolution ScreenSpot-Pro benchmark, highlighting its robustness and generality.
\end{itemize}







%% file: sec/2_related_works.tex
\section{Related Work}
\label{sec:formatting}

\noindent \textbf{GUI Agents.} 
The capability of GUI agents to automate complex tasks~\cite{mind2web, yan2023gpt, wang2024mobile} has surged with the advent of powerful Multimodal Large Language Models (MLLMs)~\cite{qwen, llama, chen2023internvl, ye2025cat+}. A cornerstone of this capability is grounding~\cite{wu2025gui, zhao2025learning}. Current approaches mainly fall into two paradigms. The first is a static full-screen paradigm, where representative systems such as SeeClick~\cite{seeclick}, OS-Atlas~\cite{wu2024atlas}, and UGround~\cite{gou2024navigating} construct large-scale annotated datasets to fine-tune MLLMs for text-to-coordinate prediction. While effective in structured layouts, this static paradigm often degrades on high-resolution, element-dense interfaces, where visual clutter disperses attention and leads to unstable localization. The limitations of static models, together with the recent rise of “think-with-image” strategies~\cite{zhu2025active, zheng2025deepeyes, cao2025ground}, have motivated a second paradigm based on iterative refinement. These methods employ a multi-step zoom-in process: ECP~\cite{lee2025training} and R-VLM~\cite{park2025r} first localize candidate regions before high-resolution refinement, DiMo-GUI~\cite{wu2025dimo} performs training-free GUI grounding via modality separation and iterative region zoom-in, and LASER~\cite{wang2025learning} is a self-evolving framework that equips VLMs with multi-step perception and adaptive region reasoning for precise instruction-relevant coordinate prediction. However, these zoom-in approaches are inherently irreversible: once the correct region is discarded early, the search cannot recover and errors accumulate. By contrast, our method employs a dynamic perception planner that can redirect and revisit regions as needed, yielding more reliable grounding in dense interfaces.

\noindent \textbf{Visual Search.} Early visual search models, inspired by human eye movements, combine saliency priors with probabilistic search or learned policies to sequentially choose fixation locations \cite{sclar2020modeling,torralba2006contextual,zhang2018finding,ye2024CAT, ye2025eyes}, but they typically operate with fixed-size attention windows and focus on reproducing scanpaths rather than accurately localizing small targets in cluttered scenes. More recent work such as SEAL \cite{wu2024vstar} couples localization modules and visual memory with large multimodal models to guide attention to promising regions, ViGoRL~\cite{sarch2025grounded} casts visual search as a sequential decision process and uses RL to learn spatially grounded fixation policies, DyFo~\cite{li2025dyfo} simulates a human-like dynamic focusing mechanism to let MLLMs zoom into fine-grained regions, and FaST~\cite{sun2024visual} controls the speed of visual reasoning via a lightweight adapter to achieve fast–slow visual search. Although these approaches have shown promise in natural-image scenarios, they do not fully account for the extreme density and structural heterogeneity of GUI layouts.
We instead formulate GUI grounding as a visual search problem and design DRS-GUI as a dynamic region-level planner that adaptively explores candidate regions under language guidance to more reliably localize fine-grained GUI targets.


\begin{figure*}[t]
    \centering
    \includegraphics[width=0.99\textwidth]{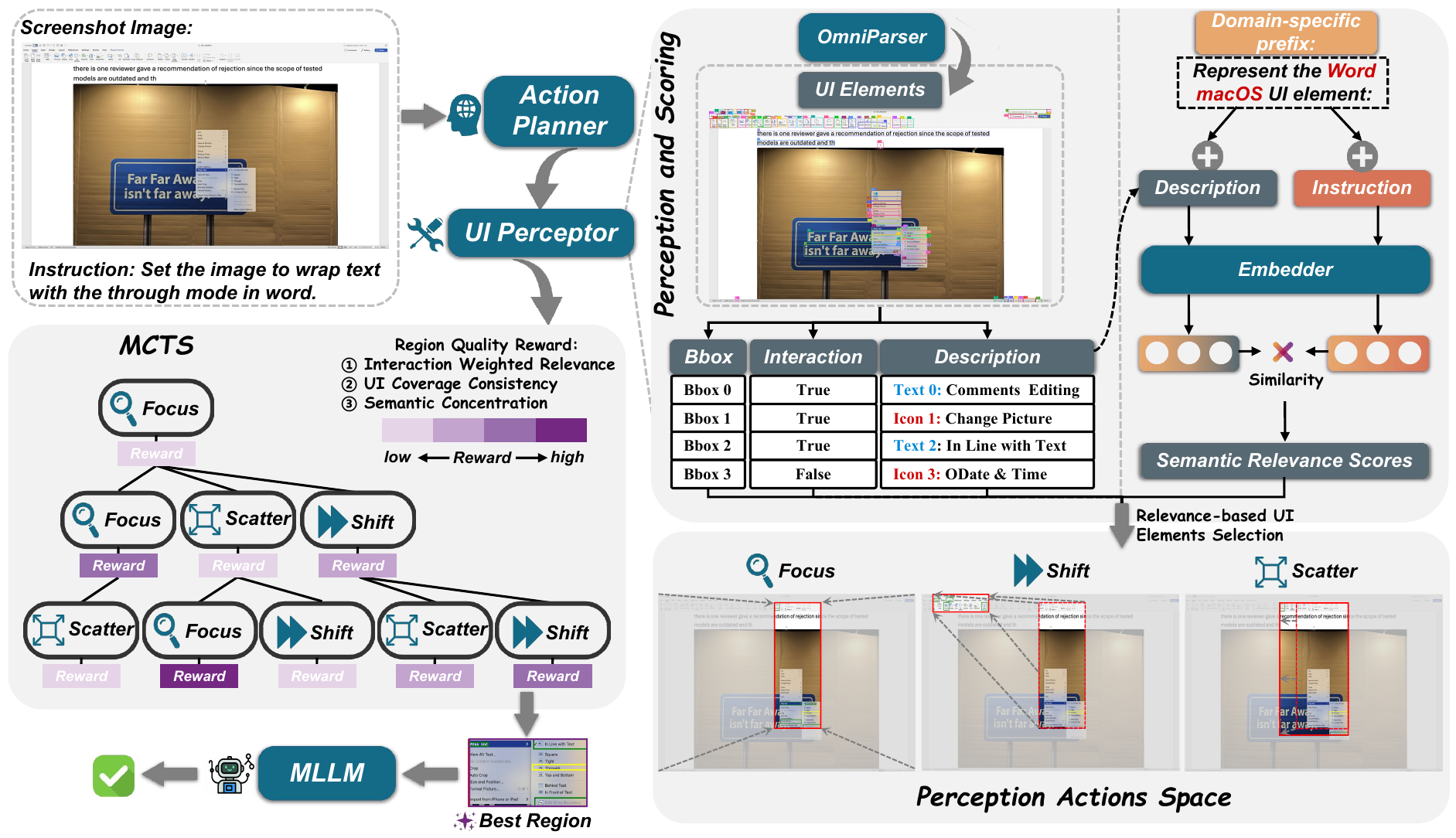}
    \vspace{-10pt}
    \caption{Processing pipeline of DRS-GUI. The UI Perceptor parses UI elements and scores their relevance to the instruction, while the Action Planner (MCTS) explores perceptual actions using region quality rewards to locate the appropriate region for final grounding.}
    \label{fig:method}
    \vspace{-10pt}
\end{figure*}


%% file: sec/3_method.tex
\section{Methodology}


We treat dynamic region search as a preparatory stage for GUI grounding, aiming to identify a reliable inference region before the base MLLM performs the final coordinate prediction. DRS-GUI adjusts the field of view by generating, evaluating, and refining candidate regions under language guidance, enabling grounding to occur only after a semantically appropriate interface region has been reached. As shown in \cref{fig:method}, DRS-GUI comprises two cooperating modules. The UI Perceptor parses the current region into structured UI elements and provides instruction-aligned semantic cues, while also executing three region-level actions (Focus, Shift, Scatter) to produce candidate views. The MCTS Action Planner then schedules these actions to construct a region search tree. Guided by a region quality reward, DRS-GUI dynamically searches and revises its perceptual scope to identify the most suitable region before passing it to the base MLLM for final coordinate prediction. This search-before-predict paradigm reduces visual redundancy and mitigates error accumulation, resulting in more stable and interpretable grounding in dense, high-resolution GUI environments.

\subsection{Problem Formulation}
The task of GUI Grounding is to map a natural language instruction $T$ to the pixel-level coordinates $p = (x, y)$ of its corresponding UI element in a screenshot $S_{full} \in \mathbb{R}^{H \times W \times 3}$. However, directly grounding on $S_{full}$ is unreliable, as the interface is high-resolution and densely filled with visually similar yet instruction-irrelevant elements.
We therefore reformulate grounding as a \emph{search-then-predict} process. A region search policy $\pi_S$ first identifies an instruction-relevant region $R_{\text{best}} \subset S_{\text{full}}$ and the base grounding model $\mathcal{M}$ performs localization only within that region:

\vspace{-1em}
\begin{equation}
R_{\text{best}} = \pi_S(S_{\text{full}}, T), \qquad
p = M(R_{\text{best}}, T)
\end{equation}

The goal of DRS-GUI is to realize  $\pi_S$ as an efficient and reversible dynamic search procedure, improving localization stability without model retraining. 

\subsection{UI Perceptor}

The UI Perceptor provides the perceptual foundation of DRS-GUI. Given a region $R \subset S_{\text{full}}$, it parses the visible interface into a structured set of UI elements and produces UI domain-aware semantic embeddings that guide the execution of perceptual actions to prioritize regions that are meaningful and semantically aligned with the user instruction.  We adopt OmniParser V2~\cite{lu2024omniparser} to extract UI elements:

\vspace{-1em}
\begin{equation}
U = \{\, u_i \mid u_i = [\, b_i, d_i, i_i \,] \,\}_{i=1}^{N}
\end{equation}

where $b_i$ denotes the bounding box of element $i$, $d_i$ denotes its semantic description (from OCR-recognized text or icon captioning), and $i_i \in \{0, 1\}$ indicates whether the element is interactive. 

To establish better semantic alignment, we generate domain-specific embeddings for both the instruction $T$ and the element descriptions $d_i$ using instructor-large embedder~\cite{su2022one}.
A domain-specific prefix $P_d$ is designed according to the application type and system type of current GUI screenshot and prepended during encoding, ensuring that both the instruction and the UI elements are embedded within a unified interaction context aligned with the active GUI environment. Concretely, we compute

\vspace{-1em}
\begin{equation}
\begin{aligned}
e_T &= \text{Embedder}(P_d \oplus T), \\
e_{d_i} &= \text{Embedder}(P_d \oplus d_i)
\end{aligned}
\end{equation}
with mean pooling over the final hidden states to obtain fixed-length vectors. The shared domain prefix places the instruction embedding $e_T$ and the element embeddings $e_{d_i}$ in a consistent semantic space. A semantic relevance score $s_i = \text{cosine}(e_T, e_{d_i})$ is then computed for each element, and these scores serve as the core cues for perceptual action execution and region reward evaluation.

Using the relevance scores list$\{s_i\}$, the UI Perceptor supports three human-like perceptual actions that dynamically adjust the spatial focus of the search region before grounding: 


\textbf{Focus.} Focus contracts the current viewing region toward UI elements that are highly relevant to the instruction. Guided by the relevance scores $\{s_i\}$, the Perceptor selects the top-$p\%$ UI elements and removes spatial outliers whose box centers deviate markedly from the cluster centroid. The remaining elements are enclosed by the minimal enclosing bounding box to form a compact crop. If the crop does not shrink sufficiently relative to the previous region, the farthest elements are iteratively pruned until a target shrink ratio is satisfied. This cluster-and-prune step concentrates semantic content while suppressing visual clutter.

\textbf{Shift.} Shift relocates the search to a different area of the interface. When high-relevance cues appear in regions that are spatially separated from the current view, Shift recenters the region around those cues. To maintain stability, these elements are grouped based on consistent layout direction (e.g., upper area, left area), and the resulting region is constructed to have minimal overlap with the previous one. Thus, Shift performs a purposeful relocation of attention—allowing the search to move out of an uninformative region and explore remaining interface space in a structured and guided manner. 

\textbf{Scatter.} Scatter expands the perceptual scope when the current region lacks strong or coherent semantic cues. It collects additional high-relevance elements outside the current region and enlarges the field of view to include them, while applying scale constraints to avoid over-expansion. This controlled expansion restores broader contextual awareness and prevents the search from becoming trapped in overly narrow local regions.



These three actions jointly support dynamic control of the perceptual scope: it tightens when semantic confidence increases, expands when additional context is needed, and shifts when more informative cues appear elsewhere. This results in a human-aligned, context-driven search behavior~\cite{torralba2006contextual, wolfe2011visual} that underlies the planner’s region search space.



\noindent
\subsection{Monte Carlo Tree Search Action Planner}  \label{sec:mcts}



The MCTS Action Planner serves as the decision-making module that schedules perceptual actions and constructs a reversible search trajectory over regions. We treat each region state $S = (R, U, \{s_i\})$ as a node in the search tree, where $R$ is the current region crop, $U$ is the parsed UI element set, and $\{s_i\}$ denotes the element-level semantic relevance scores computed by the UI Perceptor. The perceptual action $a$ space is defined as $A = \{\text{Focus}, \text{Shift}, \text{Scatter}\}$, where each action directly transforms the current region into a new candidate region, forming the edges of the perception tree. Under a fixed search budget, the planner iteratively performs the following four phases:

\noindent \textbf{Node Selection.}  
The search begins at the root node, initialized by a one-step Focus on the full image to obtain a coarse global view, ensuring that subsequent Scatter and Shift actions operate on a meaningful region. At each iteration, the planner traverses the perception tree by selecting a child node according to the Upper Confidence bounds applied to Trees (UCT) policy~\cite{kocsis2006bandit}, which balances exploitation of high-reward regions and exploration of under-visited alternatives.

\vspace{-1em}
{
\small
\begin{align}
    a^\ast = \arg\max_{a \in A(S)} \left[ Q(S, a) + c \sqrt{\frac{\ln V(S)}{V(S, a)}} \right ]
\end{align}
}
\noindent \noindent where $Q(S, a)$ estimates the expected future reward of taking action $a$ in state $S$, $V(S)$ is the visit count of $S$, and $V(S, a)$ records how many times action $a$ has been chosen from that state. The constant $c$ controls exploration, and $A(S) \subseteq A$ denotes the set of perceptual actions still available. The selection phase continues until a leaf node is reached.

\noindent \textbf{Expansion.}  
At the selected leaf node $s$, an unexecuted action $a \in A(S)$ is applied $a$ to produce a new region $R'$, forming a successor state $s'$. The new state $s'$ is inserted as a child of $S$. If all actions in $A(S)$ have been explored or the search count has reached its maximum, expansion is skipped.

\noindent \textbf{Simulation \& Backpropagation.}  
The newly proposed region is evaluated using the composite region quality reward (Sec.~\ref{sec:reward}).
Then the resulting reward is propagated up the visited path, updating both $Q$-value and visit counts $V$ for each node, allowing the search tree to accumulate guidance.

By iteratively expanding the search tree, MCTS balances exploring new regions and refining promising ones while avoiding commitment to a single perceptual path. As illustrated in \cref{fig:mcts_planning}, the planner gradually grows a tree of region states by applying perceptual actions (Focus, Shift, Scatter) and updating their values according to the region quality reward. After the search terminates, the region with the highest reward is selected and forwarded to the base MLLM for final localization:

\vspace{-1em}
{\small
\begin{align}
R_{\text{best}} = \arg\max_{R} r(R, T)
\end{align}
}
This selection mechanism restricts the base model to a compact, instruction-relevant region and improves grounding stability without any additional MLLM inference.


\noindent
\subsection{Region Quality Reward}\label{sec:reward}
\label{sec:reward}
To guide the search process, each candidate region is evaluated by a composite reward function that reflects how suitable the region is for grounding under the given instruction and whether it is worth further exploration. Given a region $R$ and the parsed UI elements $U=\{u_i\}_{i=1}^{N}$, the UI Perceptor provides instruction-conditioned relevance scores $\{s_i\}$. Using these cues, we compute three complementary terms:

\noindent\textbf{Interaction Weighted Relevance.}
Interactive elements are generally better grounding targets. Thus, relevance should emphasize such elements. We apply a simple interaction-aware weight:

\vspace{-1em}
{\small
\begin{align}
w_i = 
\begin{cases}
1, & i_i = 1 \\
\lambda, & \text{otherwise}
\end{cases}, \quad 0 < \lambda < 1
\end{align}
}
and compute a weighted relevance score:

\vspace{-1em}
{\small
\begin{align}
r_{\text{rel}} = \frac{\sum_{i=1}^{N} w_i \cdot s_i}{\sum_{i=1}^{N} w_i + \varepsilon}
\end{align}
}
This encourages regions where high-relevance cues correspond to interactive targets, suppressing decorative or static text.

\noindent\textbf{UI Coverage Consistency.}
Grounding is more reliable when the perceptual region contains actual UI structure rather than blank background or decorative space. We quantify the proportion of the region occupied by detected UI elements:

\vspace{-1em}
{\small
\begin{align}
r_{\text{cov}} = \frac{\sum_{i=1}^{N} \text{Area}(b_i)}{\text{Area}(R)}
\end{align}
}
\vspace{-1em}

Higher coverage indicates that the region contains meaningful interface components, while lower coverage suggests visually redundant or non-informative areas.

\noindent\textbf{Semantic Concentration.}
We assess whether relevance within the crop is focused or diffuse. Using \emph{all} $N$ elements, we normalize scores with temperature $\tau$:

{\small
\begin{equation}
p_i = \frac{\exp(s_i/\tau)}{\sum_{j=1}^{N}\exp(s_j/\tau)}
\end{equation}
}
and compute normalized entropy over the full set:

{\small
\begin{equation}
r_{\text{con}} = 1 - \frac{-\sum_{i=1}^{N} p_i \log p_i}{\log(N + \varepsilon)}
\end{equation}
}
Higher $r_{\text{con}}$ indicates a clear semantic focus suitable for grounding.
\input{table/screenspot_pro}

\noindent\textbf{Final Reward.}
The overall region reward is defined as a weighted combination of the three components:

\vspace{-1em}
{\small
\begin{align}
r(R, T) = \alpha \cdot r_{\text{rel}} + \beta \cdot r_{\text{cov}} + \gamma \cdot r_{\text{con}}
\end{align}
}
where $\alpha, \beta, \gamma$ control the relative contributions of the three reward terms. This reward guides the planner toward regions that are semantically meaningful, operationally relevant, and visually informative, enabling robust grounding.


%% file: table/screenspot_pro.tex
\begin{table*}[!ht]
\setlength{\tabcolsep}{3.5pt} %
\centering
\caption{Comparison of various models on ScreenSpot-Pro~\cite{li2025screenspot}. Without any extra training or external data, DRS-GUI substantially improves the grounding performance of Qwen2.5-VL-3\&7B~\cite{bai2025qwen2} and UGround-V1-2\&7B~\cite{gou2024navigating}.}
\resizebox{\linewidth}{!}{
\begin{tabular}{l|ccc|ccc|ccc|ccc|ccc|ccc|ccc}
\toprule
\textbf{Agent Model} & \multicolumn{3}{c|}{\textbf{Development}} & \multicolumn{3}{c|}{\textbf{Creative}} & \multicolumn{3}{c|}{\textbf{CAD}} & \multicolumn{3}{c|}{\textbf{Scientific}} & \multicolumn{3}{c|}{\textbf{Office}} & \multicolumn{3}{c|}{\textbf{OS}} & \multicolumn{3}{c}{\textbf{Avg}} \\
\cmidrule(lr){2-4} \cmidrule(lr){5-7} \cmidrule(lr){8-10} \cmidrule(lr){11-13} \cmidrule(lr){14-16} \cmidrule(lr){17-19} \cmidrule(lr){20-22}
 & \textbf{text} & \textbf{icon} & \textbf{avg} & \textbf{text} & \textbf{icon} & \textbf{avg} & \textbf{text} & \textbf{icon} & \textbf{avg} & \textbf{text} & \textbf{icon} & \textbf{avg} & \textbf{text} & \textbf{icon} & \textbf{avg} & \textbf{text} & \textbf{icon} & \textbf{avg} & \textbf{text} & \textbf{icon} & \textbf{avg} \\
\midrule
QwenVL-7B~\cite{Qwen-VL} & 0.0 & 0.0 & 0.0 & 0.0 & 0.0 & 0.0 & 0.0 & 0.0 & 0.0 & 0.7 & 0.0 & 0.4 & 0.0 & 0.0 & 0.0 & 0.0 & 0.0 & 0.0 & 0.1 & 0.0 & 0.1 \\
GPT-4o~\cite{gpt4} & 1.3 & 0.0 & 0.7 & 1.0 & 0.0 & 0.6 & 2.0 & 0.0 & 1.5 & 2.1 & 0.0 & 1.2 & 1.1 & 0.0 & 0.6 & 0.0 & 0.0 & 0.0 & 1.3 & 0.0 & 0.8 \\
SeeClick~\cite{seeclick} & 0.6 & 0.0 & 0.3 & 1.0 & 0.0 & 0.6 & 2.5 & 0.0 & 1.9 & 3.5 & 0.0 & 2.0 & 1.1 & 0.0 & 0.5 & 2.8 & 0.0 & 1.5 & 1.8 & 0.0 & 1.1 \\
Qwen2-VL-7B~\cite{Qwen2VL} & 2.6 & 0.0 & 1.3 & 1.5 & 0.0 & 0.9 & 0.5 & 0.0 & 0.4 & 6.3 & 0.0 & 3.5 & 3.4 & 1.9 & 3.0 & 0.9 & 0.0 & 0.5 & 2.5 & 0.2 & 1.6 \\
OS-Atlas-4B~\cite{wu2024atlas} & 7.1 & 0.0 & 3.7 & 3.0 & 1.4 & 2.3 & 2.0 & 0.0 & 1.5 & 9.0 & 5.5 & 7.5 & 5.1 & 3.8 & 4.4 & 5.6 & 0.0 & 3.1 & 5.0 & 1.7 & 3.7 \\
ShowUI-2B\cite{lin2024showui} & 16.9 & 1.4 & 9.4 & 9.1 & 0.0 & 5.3 & 2.5 & 0.0 & 1.9 & 13.2 & 7.3 & 10.6 & 15.3 & 7.5 & 13.5 & 10.3 & 2.2 & 6.6 & 10.8 & 2.6 & 7.7 \\
CogAgent-18B ~\cite{cogagent}& 14.9 & 0.7 & 8.0 & 9.6 & 0.0 & 5.6 & 7.1 & 3.1 & 6.1 & 22.2 & 1.8 & 13.4 & 13.0 & 0.0 & 6.5 & 5.6 & 0.0 & 3.1 & 12.0 & 0.8 & 7.7 \\
Aria-UI~\cite{yang2025aria} & 16.2 & 0.0 & 8.4 & 23.7 & 2.1 & 14.7 & 7.6 & 1.6 & 6.1 & 27.1 & 6.4 & 18.1 & 20.3 & 1.9 & 16.1 & 4.7 & 0.0 & 2.6 & 17.1 & 2.0 & 11.3 \\
UGround-7B~\cite{gou2024navigating} & 26.6 & 2.1 & 14.7 & 27.3 & 2.8 & 17.0 & 14.2 & 1.6 & 11.1 & 31.9 & 2.7 & 19.3 & 31.6 & 11.3 & 27.9 & 17.8 & 0.0 & 9.7 & 25.0 & 2.8 & 16.5 \\
Claude Comp.Use~\cite{hu2024dawn} & 22.0 & 3.9 & 12.6 & 25.9 & 3.4 & 16.8 & 14.5 & 3.7 & 11.9 & 33.9 & 15.8 & 25.8 & 30.1 & 16.3 & 26.2 & 11.0 & 4.5 & 8.1 & 23.4 & 7.1 & 17.1 \\
OS-Atlas-7B~\cite{wu2024atlas} & 33.1 & 1.4 & 17.7 & 28.8 & 2.8 & 17.9 & 12.2 & 4.7 & 10.3 & 37.5 & 7.3 & 24.4 & 33.9 & 5.7 & 27.4 & 27.1 & 4.5 & 16.8 & 28.1 & 4.0 & 18.9 \\
UI-TARS-7B~\cite{qin2025ui} & 58.4 & 12.4 & 36.1 & 50.0 & 9.1 & 32.8 & 20.8 & 9.4 & 18.0 & 63.9 & 31.8 & 50.0 & 63.3 & 20.8 & 53.5 & 30.8 & 16.9 & 24.5 & 47.8 & 16.2 & 35.7 \\
\midrule

Qwen2.5-VL-3B~\cite{bai2025qwen2}  & 22.1 & 1.4 & 12.0 & 26.8 & 2.1  & 16.4 & 9.1 & 7.3 &  7.3 & 38.2 & 7.3 & 24.8 & 33.9 & 15.1 & 29.6 & 10.3 & 1.1 & 6.1 & 23.6 & 3.8 & 16.1 \\

+ \emph{DRS-GUI} & \textbf{44.8} & \textbf{6.9} & \textbf{26.4} & \textbf{44.9} & \textbf{7.7} & \textbf{29.3} & \textbf{21.3} & \textbf{9.4} & \textbf{18.4} & \textbf{50.0} & \textbf{11.8} & \textbf{33.5} & \textbf{52.0} & \textbf{24.5} & \textbf{45.7} & \textbf{26.2} & \textbf{10.1} & \textbf{18.9} & \textbf{40.1} & \textbf{10.3} & \textbf{28.7}
\\

Qwen2.5-VL-7B~\cite{bai2025qwen2} & 46.8 & 4.1 &  26.1 & 35.9 & 7.7  & 24.0 & 16.8 & 1.6 & 13.0  & 49.3 & 7.3 & 31.1 & 52.5 & 20.8 & 45.2 & 37.4 & 6.7 & 23.5 & 38.9 & 7.1 & 26.8 \\

+ \emph{DRS-GUI} & \textbf{60.4} & \textbf{13.1} & \textbf{37.5} & \textbf{54.6} & \textbf{16.8} & \textbf{38.7} & \textbf{42.1} & \textbf{9.4} & \textbf{34.1} & \textbf{59.7} & \textbf{16.4} & \textbf{41.0} & \textbf{66.7} & \textbf{26.4} & \textbf{57.4} & \textbf{52.3} & \textbf{23.6} & \textbf{39.3} & \textbf{55.7} & \textbf{16.9} & \textbf{40.9}
\\

\midrule
UGround-V1-2B~\cite{gou2024navigating} & 48.7 & 4.1 & 27.1 & 43.4 & 4.8 & 27.3 & 17.8 & 6.3 & 14.9 & 52.1 & 15.5 & 36.2 & 44.6 & 13.2 & 37.4 & 27.1 & 4.5 & 16.8 & 38.7 & 7.5 & 26.8 \\

+ \emph{DRS-GUI} & \textbf{62.3} & \textbf{11.7} & \textbf{37.8} & \textbf{49.5} & \textbf{12.6} & \textbf{34.0} & \textbf{32.5} & \textbf{12.5} & \textbf{27.6} & \textbf{56.3} & \textbf{20.0} & \textbf{40.6} & \textbf{62.7} & \textbf{28.3} & \textbf{54.8} & \textbf{52.3} & \textbf{22.5} & \textbf{38.8} & \textbf{51.8} & \textbf{16.6} & \textbf{38.3}
\\

UGround-V1-7B~\cite{gou2024navigating} & 51.9 & 3.4 & 28.4 & 48.0 & 9.1 & 31.7 & 20.0 & 1.6 & 15.3 & 57.6 & 16.4 & 39.8 & 61.6 & 13.2 & 50.4 & 37.4 & 7.9 & 25.0 & 45.6 & 8.4 & 31.4 \\
+ \emph{DRS-GUI} & \textbf{66.9} & \textbf{21.4} & \textbf{44.8} & \textbf{56.6} & \textbf{16.1} & \textbf{39.6} & \textbf{43.7} & \textbf{15.6} & \textbf{36.8} & \textbf{73.0} & \textbf{25.5} & \textbf{52.4} & \textbf{73.4} & \textbf{30.2} & \textbf{63.5} & \textbf{52.3} & \textbf{25.8} & \textbf{40.3} & \textbf{60.6} & \textbf{21.7} & \textbf{45.7}
\\

\bottomrule
\end{tabular}}
\vspace{-0.1in}
\label{tab:screenspot_pro_results}
\end{table*}

%% file: sec/4_experiments.tex
\section{Experiments}
\label{sec:experiments}



\subsection{Experimental Settings}

\noindent\textbf{Datasets \& Evaluation Metrics}
We evaluate DRS-GUI on three widely used GUI grounding benchmarks: ScreenSpot-V1~\cite{seeclick}, ScreenSpot-v2~\cite{wu2024atlas}, and ScreenSpotPro~\cite{li2025screenspot}. ScreenSpot-V1 and ScreenSpot-v2 cover general GUI domains across Mobile, Web, and Desktop interfaces. ScreenSpot-Pro focuses on high-resolution professional applications, featuring 23 real software interfaces with denser and more complex UI layouts, making grounding substantially more challenging. Following standard practice, a grounding prediction is considered correct if the predicted point lies within the ground-truth bounding box~\cite{seeclick}. We report grounding accuracy as the primary evaluation metric across all experiments.

\noindent\textbf{Baseline Methods} 
To demonstrate the generality and plug-and-play nature of DRS-GUI, we evaluate it on both general open-source MLLMs and GUI-specific models, covering small and large parameter scales. For general-purpose MLLMs, we adopt Qwen2.5-VL-3B-Instruct and Qwen2.5-VL-7B-Instruct~\cite{qwen2.5}. For GUI-specific models, we experiment with UGround-V1-2B and UGround-V1-7B~\cite{gou2024navigating} which are explicitly trained on GUI grounding dataset. These models allow us to assess the robustness and effectiveness of DRS-GUI as a training-free enhancement to GUI agents.

\input{table/screenspot}
\input{table/screenspotv2}

\begin{figure}[ht]
    \centering
    \includegraphics[width=0.9\linewidth]{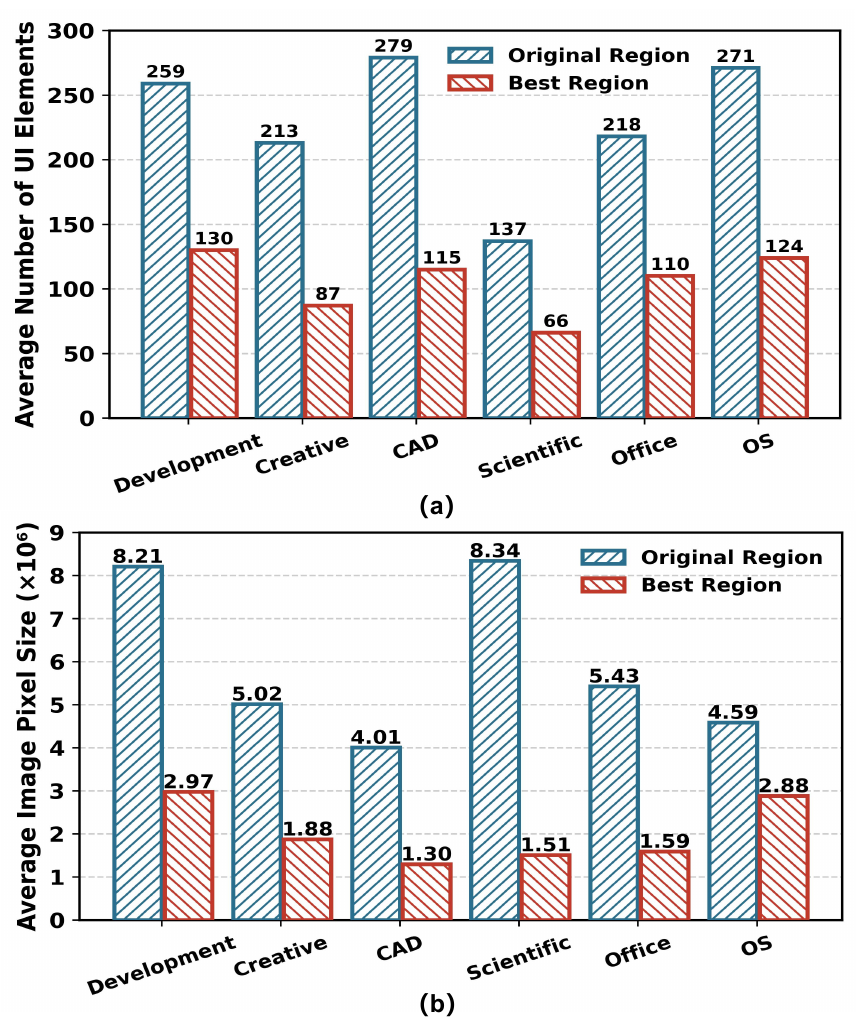}
    \vspace{-10pt}
    \caption{Redundancy reduction of our dynamic region search across application categories.
(a) Average image pixel size and (b) average number of UI elements for Original vs. Bst regions; the focused crop cuts input area and UI elements by 64\% and 54\% on average. }
    \label{fig:Redundancy-analysis}
    \vspace{-10pt}
\end{figure}

\begin{figure*}[t]
    \centering
    \includegraphics[width=0.9\textwidth]{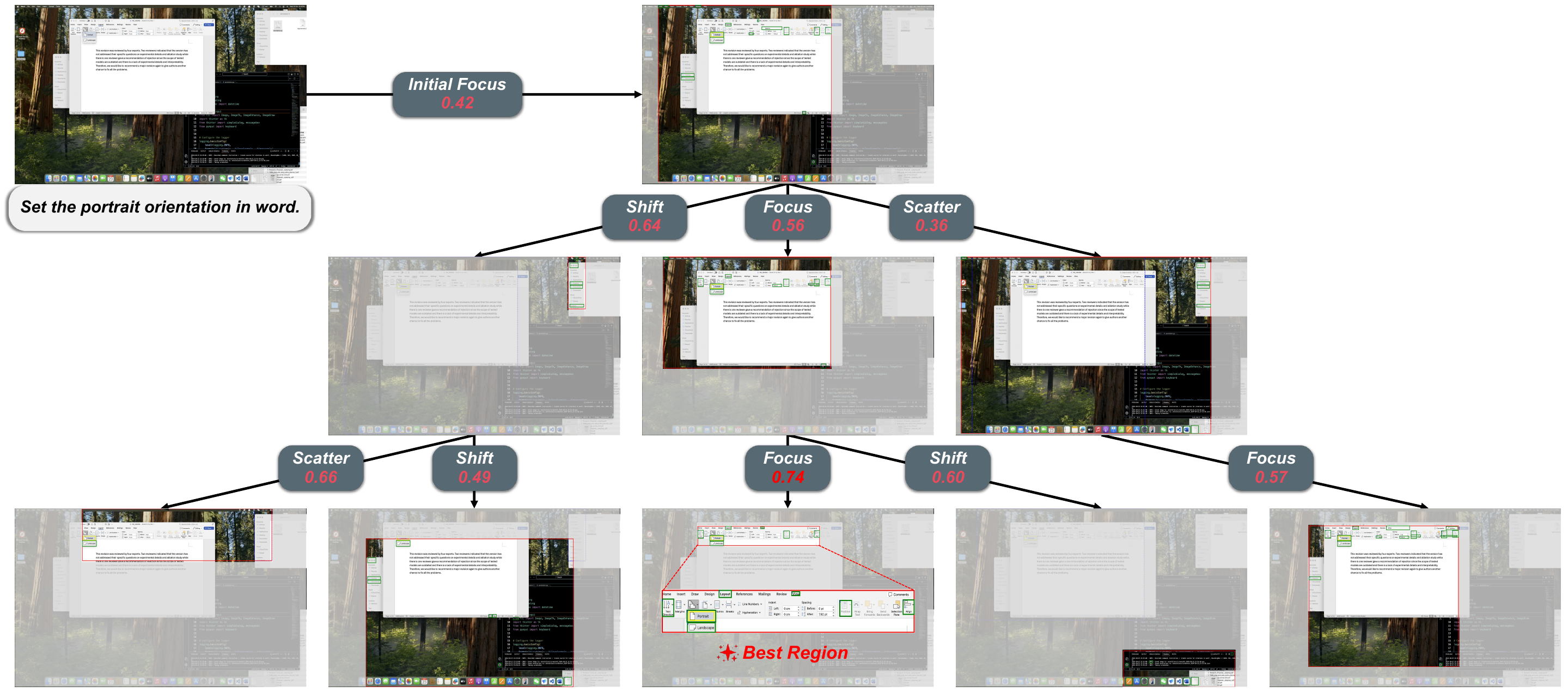}
    \vspace{-10pt}
    \caption{Visualization of the MCTS-based perceptual planning process. The tree illustrates how the planner dynamically adjusts perceptual actions to progressively localize the optimal region. The top node shows the Initial Focus and its reward. Guided by the UCT policy, the planner expands nodes by selecting one of the three actions, each producing a candidate region evaluated by the region quality reward. Through iterative exploration and backpropagation, the search converges to the Best Region}
    \label{fig:mcts_planning}
    \vspace{-10pt}
\end{figure*}

\begin{figure}[t]
    \centering
    \includegraphics[width=\linewidth]{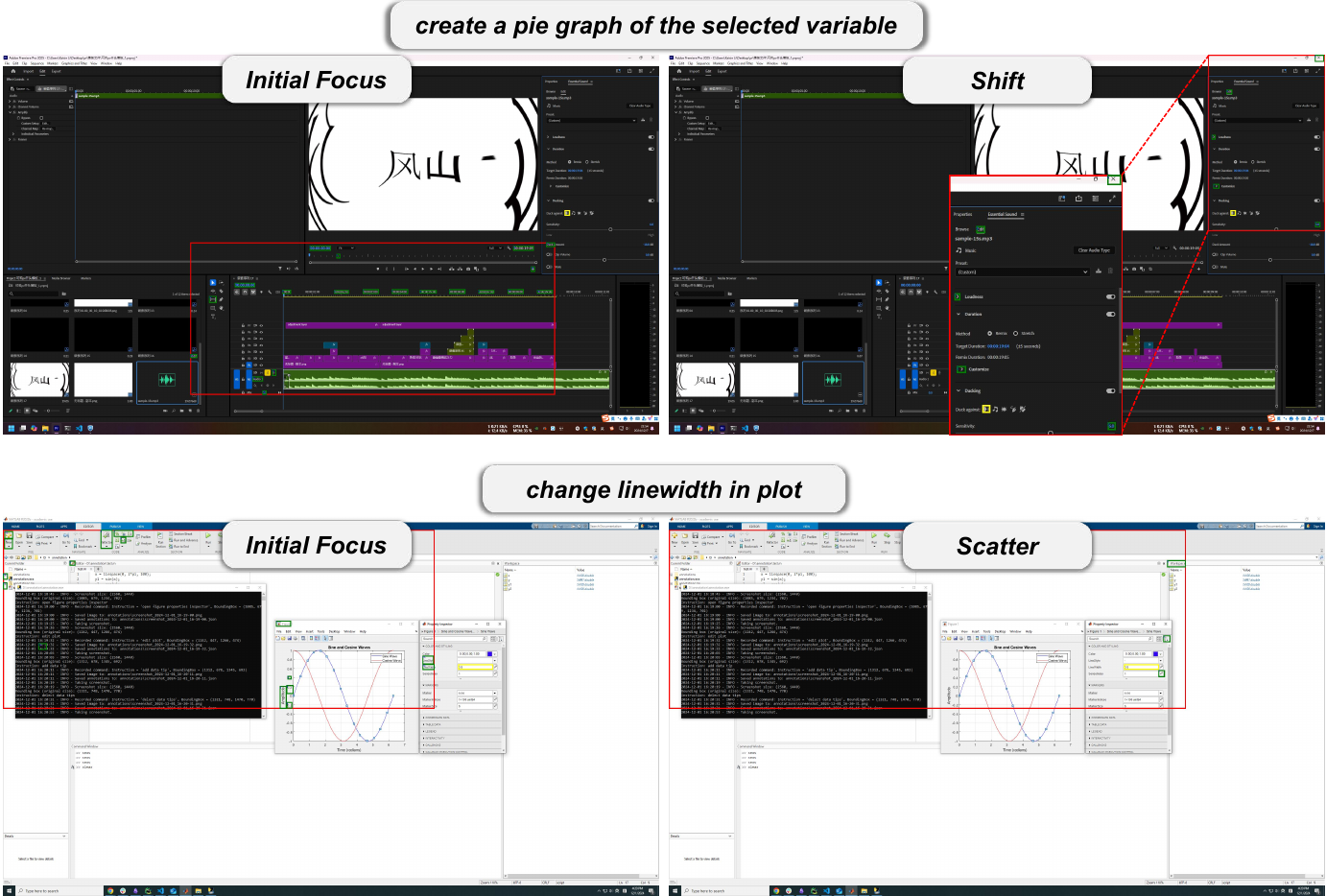}
    \caption{Qualitative examples of perceptual actions. The left red boxes denote the initial Focus regions, while the right red boxes represent the final best regions after applying the corresponding perceptual actions. Yellow boxes indicate the ground-truth UI elements, and green dots mark the final predicted coordinates.}
    \label{fig:qualitative_actions}
    \vspace{-10pt}
\end{figure}




\noindent\textbf{Implementation Details}  
All open-weight models (OmniParser V2 and Instructor-large) are used directly without finetuning. In practice, Focus keeps the top 15\% relevant elements; Scatter adds the top 10\% external elements while limiting area expansion to $1.5\times$; and Shift re-centers on the top 15\% external anchors with $\text{IoU}\le0.3$ to avoid overlap. The MCTS planner runs with a rollout budget of $N=8$, depth limit $H=3$, and UCT exploration constant $c=1$. For region evaluation, we use the composite reward described in Sec.~\ref{sec:reward}, with weights $\alpha=0.4$, $\beta=0.4$, and $\gamma=0.2$, and further analysis of these hyper-parameters is provided in supplementary materials. All experiments are conducted with two NVIDIA A6000 GPUs.


\subsection{Main Results}
\begin{figure}[ht]
    \centering
    \includegraphics[width=\linewidth]{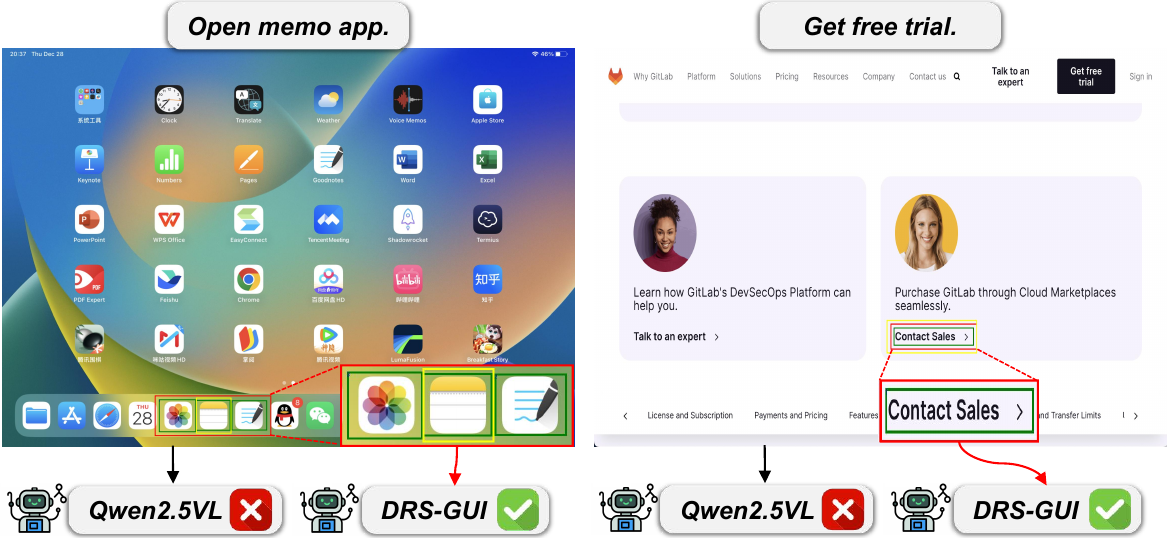}
    \caption{Comparison between the responses of Qwen2.5-VL and our method DRS-GUI on ScreenSpot-v2 cases. The final best region is highlighted in the image using red bounding boxes.}
    \label{fig:v2case}
    \vspace{-6pt}
\end{figure}

\noindent\textbf{ScreenSpot-V1\&V2} We evaluate DRS-GUI on ScreenSpot and ScreenSpot-V2 by integrating it into both Qwen2.5-VL-7B and UGround-V1-7B. As shown in  \cref{table:Mind2Web-benchmark} and \cref{table:screenspotv2}, DRS-GUI consistently improves grounding accuracy across all settings. 
On ScreenSpot-V2, it boosts Qwen2.5-VL-7B by about 4\% and UGround-V1-7B by roughly 4.2\%. Comparable improvements are observed on ScreenSpot-V1 for both models. Notably, the improvements are more pronounced in icon and widget cases, where visual ambiguity is higher, indicating that dynamic region search effectively suppresses irrelevant UI elemens and stabilizes grounding behavior. Representative qualitative cases are shown in \cref{fig:v2case}.

\begin{figure}[ht]
    \centering
    \includegraphics[width=\linewidth]{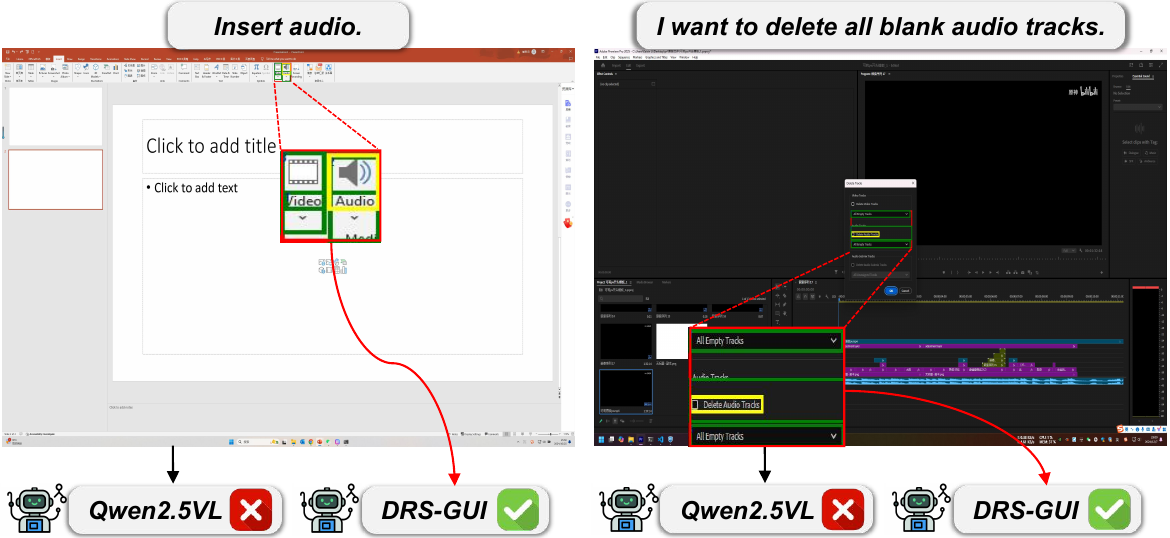}
    \caption{Comparison between the responses of Qwen2.5-VL and our method DRS-GUI on ScreenSpot-Pro cases. The final best region is highlighted in the image using red bounding boxes. }
    \label{fig:procase}
    \vspace{-10pt}
\end{figure}

\noindent\textbf{ScreenSpot-Pro}
We further validate our approach on the challenging ScreenSpot-Pro benchmark (\cref{tab:screenspot_pro_results}). Across all six high-resolution application domains, our training-free framework yields substantial gains: accuracy improves by 14.1\% on Qwen2.5-VL-7B and 14.3\% on UGround-V1-7B. Notably, Qwen2.5-VL-3B + DRS-GUI (28.7\%) even surpasses larger specialized models such as OS-Atlas-7B (18.9\%), showing that effective grounding depends more on adaptive perceptual search than on model size. The redundancy-reduction analysis (\cref{fig:Redundancy-analysis}) further confirms that DRS-GUI consistently narrows the search space by focusing on compact, instruction-relevant regions, enabling models to maintain precise grounding under clutter-heavy professional interfaces. Representative qualitative cases are shown in \cref{fig:procase}.

\input{table/actionablation}
\input{table/rewardablation}

\subsection{Ablation Study}
\noindent\textbf{Effect of Perceptual Actions Space} We further analyze the effect of perceptual action design on grounding performance. As shown in \cref{table:actionab}, introducing the Focus action alone improves accuracy by 2.2\%, confirming its key role in refining visual attention. Combining Focus and Shift further boosts performance to 91.0\%, though a slight drop appears on the Mobile subset due to occasional over-shifting of compact layouts. Adding Scatter achieves the best result of 91.8\%, indicating that a full action space yields the most balanced and robust perceptual search behavior.

To further illustrate the role of different perceptual actions, \cref{fig:qualitative_actions} presents two representative cases. In the first case, the initial Focus region fails to attend to the area containing the ground-truth target, while the final best region produced through a Shift operation accurately aligns with it, demonstrating the benefit of spatial relocation. In the second case, the Scatter action effectively broadens contextual coverage and captures the true target that Focus initially missed. These qualitative results visually confirm that both Shift and Scatter complement Focus by adaptively refining and expanding perception when initial attention is misplaced.

\begin{figure}[t]
    \centering
    \includegraphics[width=0.8\linewidth]{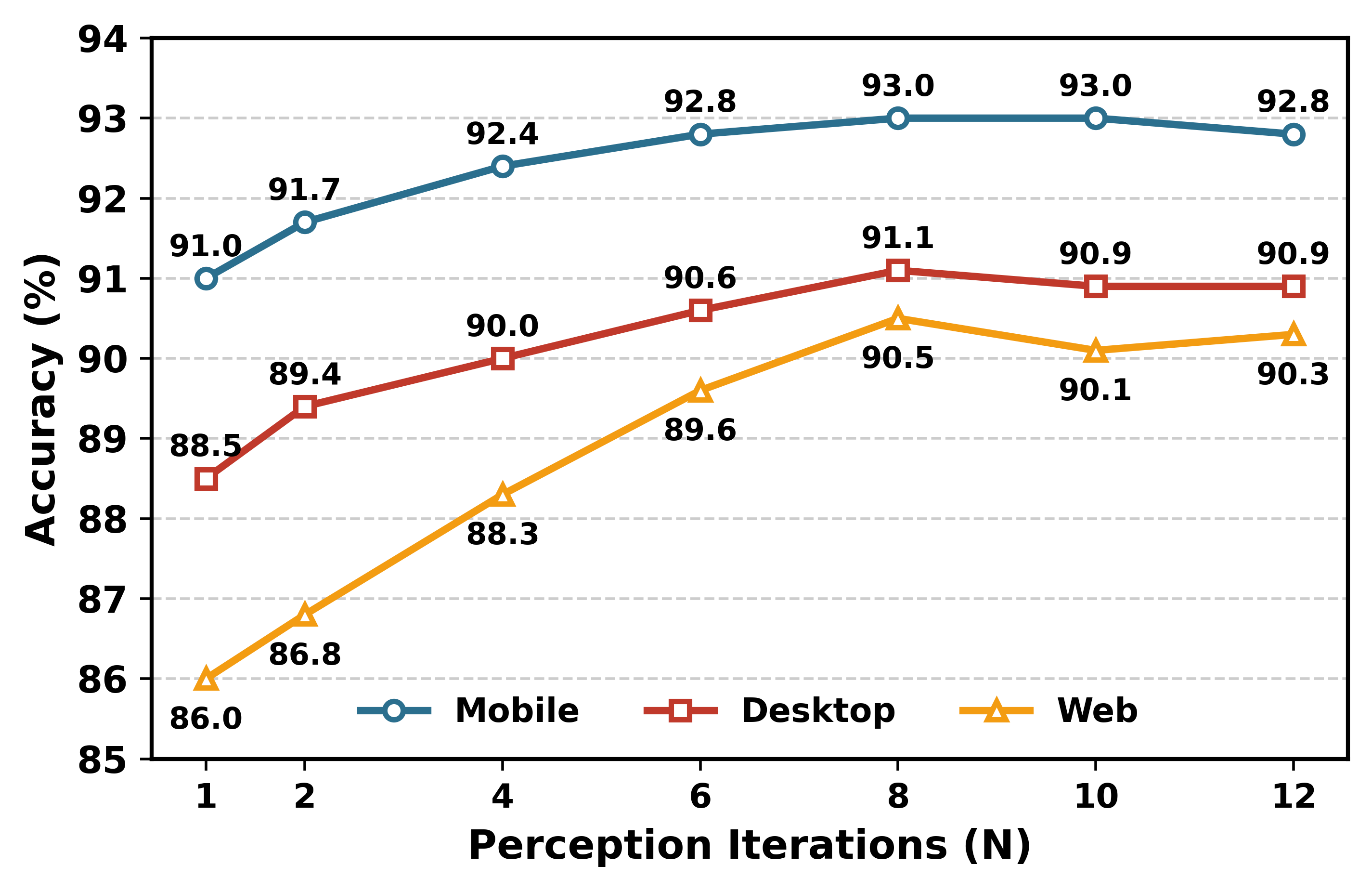}
    \caption{Accuracy under different perception iteration counts $N$ across Mobile, Desktop, and Web interfaces.}
    \label{fig:perception_iterations}
    \vspace{-10pt}
\end{figure}

\noindent\textbf{Effect of Region Quality Reward}
We next analyze how different region quality rewards influence grounding accuracy. As shown in \cref{table:rewardab}, introducing the Interactive Weighted Relevance term alone yields a 2.0\% gain, highlighting its importance in guiding attention toward semantically meaningful and actionable areas. Adding the UI Coverage Consistency reward provides slight additional improvement by encouraging spatial completeness, though overexpansion can occasionally introduce redundancy and cause small drops on Web layouts. Incorporating the Semantic Concentration reward  achieves the best overall accuracy of 91.8\%, notably improving performance on complex Web scenes. This shows that promoting semantic compactness helps the model maintain focus and stability, complementing the effects of relevance and coverage for balanced region evaluation.

\noindent\textbf{Perception Iteration Budget}
To investigate the influence of the iteration budget, we varied $N$ while fixing the maximum tree depth at 3. As shown in \cref{fig:perception_iterations}, accuracy consistently improves as $N$ increases, but the performance gain becomes marginal beyond $N=8$ iterations, indicating diminishing returns as additional exploration introduces redundant regional evaluations. Consequently, we select $N=8$ as the default iteration budget, as it achieves a strong balance between accuracy and computational efficiency.


%% file: table/screenspot.tex
\begin{table}[ht]
    \centering
    \renewcommand{\arraystretch}{1.} 
    \setlength{\tabcolsep}{0.07cm}
    \caption{Comparison of various models on ScreenSpot-V1~\cite{seeclick}.}
    \resizebox{1.0\linewidth}{!}{
    \begin{tabular}{@{}lcccccccc@{}}
        \toprule
        \multirow{2}{*}{\textbf{Method}}  &  \multicolumn{2}{c}{\textbf{Mobile}} & \multicolumn{2}{c}{\textbf{Desktop}} & \multicolumn{2}{c}{\textbf{Web}} & \multirow{2}{*}{\textbf{Avg.}} \\ 
        \cline{2-3} \cline{4-5} \cline{6-7}
         &{Text}&{Icon/Widget}&{Text}&{Icon/Widget}&{Text}&{Icon/Widget}& \\ 
        \midrule
        Llama 3.2-V-11B~\cite{grattafiori2024llama}&14.7&5.7&9.3&4.3&4.3&4.4&7.1\\
        GPT-4V\cite{gpt4v} & 22.6 & 24.5 & 20.2 & 11.8 & 9.2 & 8.8 & 16.2 \\
        Fuyu-8B~\cite{fuyu-8b} & 41.0 & 1.3 & 33.0 & 3.6 & 33.9 & 4.4 & 19.5 \\

        InternVL2-8B~\cite{chen2024far}  & 74.0 & 25.8 & 54.6 & 27.1 & 38.3 & 31.6 & 41.9 \\
        CogAgent-18B~\cite{cogagent} & 67.0 & 24.0 & 74.2 & 20.0 & 70.4 & 28.6 & 47.4 \\
        SeeClick~\cite{seeclick}  & 78.0 & 52.0 & 72.2 & 30.0 & 55.7 & 32.5 & 53.4 \\
        \hline
        Qwen2.5-VL-7B~\cite{bai2025qwen2}   & 96.7 & 77.3 & 89.2 & 70.7 & 88.7 & 79.1 & 84.9 \\
+ \emph{DRS-GUI}     & \textbf{97.8} & \textbf{81.7} & \textbf{94.3} & \textbf{77.1} & \textbf{92.2} & \textbf{84.5} & \textbf{88.9} \\
UGround-V1-7B~\cite{gou2024navigating}   & 93.0 & 79.9 & 93.8 & 76.4 & 90.9 & 84.0 & 86.3 \\
+ \emph{DRS-GUI}     & \textbf{94.2} & \textbf{84.7} & \textbf{95.9} & \textbf{82.3} & \textbf{93.5} & \textbf{86.9} & \textbf{89.9} \\
        \bottomrule
    \end{tabular}
    }
    \vspace{-10pt}
    \label{table:Mind2Web-benchmark}
\end{table}

%% file: table/screenspotv2.tex
\begin{table}[ht]
    \centering
    \renewcommand{\arraystretch}{1.} 
    \setlength{\tabcolsep}{0.07cm}
    \caption{Comparison of various models on ScreenSpot-V2~\cite{wu2024atlas}.}
    \resizebox{1.0\linewidth}{!}{
    \begin{tabular}{@{}lcccccccc@{}}
        \toprule
        \multirow{2}{*}{\textbf{Method}}  &  \multicolumn{2}{c}{\textbf{Mobile}} & \multicolumn{2}{c}{\textbf{Desktop}} & \multicolumn{2}{c}{\textbf{Web}} & \multirow{2}{*}{\textbf{Avg.}} \\ 
        \cline{2-3} \cline{4-5} \cline{6-7}
         &{Text}&{Icon/Widget}&{Text}&{Icon/Widget}&{Text}&{Icon/Widget}& \\ 
        \midrule

    InternVL2-4B~\cite{chen2024far} & 9.2 & 4.8 & 4.6 & 4.3 & 0.9 & 0.1 & 4.3 \\
    Fuyu-8B~\cite{fuyu-8b}           & 41.0 & 1.3 & 33.0 & 3.6 & 33.9 & 4.4 & 19.5 \\
    Qwen2-VL-7B~\cite{wang2024qwen2}    & 61.3 & 39.3 & 52.0 & 45.0 & 33.0 & 21.8 & 42.9 \\
    CogAgent-18B~\cite{cogagent}      & 67.0 & 24.0 & 74.2 & 20.0 & 70.4 & 28.6 & 47.4 \\
    SeeClick~\cite{seeclick}       & 78.0 & 52.0 & 72.2 & 30.0 & 55.7 & 32.5 & 53.4 \\ 
    OS-Atlas-4B~\cite{wu2024atlas}    & 85.7 & 58.5 & 72.2 & 45.7 & 82.6 & 63.1 & 70.1 \\
    UGround-7B~\cite{gou2024navigating}     & 82.8 & 60.3 & 82.5 & 63.6 & 80.4 & 70.4 & 73.3 \\


        \hline

Qwen2.5-VL-7B~\cite{bai2025qwen2}   & 99.0 & 84.4 & 87.6 & 65.7 & 90.2 & 79.8 & 86.5 \\
+ \emph{DRS-GUI}     & \textbf{99.1} & \textbf{87.7} & \textbf{92.8} & \textbf{74.3} & \textbf{96.6} & \textbf{83.3} & \textbf{90.5} \\
UGround-V1-7B~\cite{gou2024navigating}   & 95.0 & 83.3 & \textbf{95.0} & 77.8 & 92.1 & 77.2 & 87.6 \\
+ \emph{DRS-GUI}     & \textbf{96.6} & \textbf{88.2} & 94.9 & \textbf{84.3} & \textbf{94.0} & \textbf{87.7} & \textbf{91.8} \\
        \bottomrule
    \end{tabular}
    }
    \vspace{-10pt}
    \label{table:screenspotv2}
\end{table}

%% file: table/actionablation.tex
\begin{table}[ht]
\centering
\caption{Ablation study on the effect of different perceptual action combinations on ScreenSpot-V2 using UGround-V1-7B}
\resizebox{0.9\linewidth}{!}{%
\begin{tabular}{ccccccc}
\toprule
\textbf{Focus} & \textbf{Shift} & \textbf{Scatter} & \textbf{Mobile} & \textbf{Desktop} & \textbf{Web} & \textbf{Avg} \\
\midrule
 &  &  & 90.0 & 87.8 & 85.2 & 87.6 \\
\checkmark &  &  & 92.0\textsubscript{\textcolor{TealBlue}{$\uparrow$\,2.0}} & 89.8\textsubscript{\textcolor{TealBlue}{$\uparrow$\,2.0}} & 87.4\textsubscript{\textcolor{TealBlue}{$\uparrow$\,2.2}} & 89.8\textsubscript{\textcolor{TealBlue}{$\uparrow$\,2.2}} \\
\checkmark & \checkmark &  & 91.4\textsubscript{\textcolor{BrickRed}{$\downarrow$\,0.6}} & 91.0\textsubscript{\textcolor{TealBlue}{$\uparrow$\,1.2}} & 90.4\textsubscript{\textcolor{TealBlue}{$\uparrow$\,3.0}} & 91.0\textsubscript{\textcolor{TealBlue}{$\uparrow$\,1.2}} \\
\checkmark & \checkmark & \checkmark & \textbf{93.0}\textsubscript{\textcolor{TealBlue}{$\uparrow$\,1.6}} & \textbf{91.1}\textsubscript{\textcolor{TealBlue}{$\uparrow$\,0.1}} & \textbf{90.5}\textsubscript{\textcolor{TealBlue}{$\uparrow$\,0.1}} & \textbf{91.8}\textsubscript{\textcolor{TealBlue}{$\uparrow$\,0.8}} \\
\bottomrule
\end{tabular}%
}
\vspace{-10pt}
\label{table:actionab}
\end{table}

%% file: table/rewardablation.tex
\begin{table}[ht]
\centering
\caption{Ablation study on the effect of different region quality  reward combinations on ScreenSpot-V2 using UGround-V1-7B.}
\resizebox{0.9\linewidth}{!}{%
\begin{tabular}{ccccccc}
\toprule
\textbf{$r_{rel}$} & \textbf{$r_{cov}$} & \textbf{$r_{con}$} & \textbf{Mobile} & \textbf{Desktop} & \textbf{Web} & \textbf{Avg} \\
\midrule
 &  &  & 90.0 & 87.8 & 85.2 & 87.6 \\
\checkmark &  &  & 91.8\textsubscript{\textcolor{TealBlue}{$\uparrow$\,1.8}} & 89.5\textsubscript{\textcolor{TealBlue}{$\uparrow$\,1.7}} & 87.2\textsubscript{\textcolor{TealBlue}{$\uparrow$\,2.0}} & 89.6\textsubscript{\textcolor{TealBlue}{$\uparrow$\,2.0}} \\
\checkmark & \checkmark &  & 92.2\textsubscript{\textcolor{TealBlue}{$\uparrow$\,0.4}} & 90.1\textsubscript{\textcolor{TealBlue}{$\uparrow$\,0.6}} & 87.0\textsubscript{\textcolor{BrickRed}{$\downarrow$\,0.2}} & 89.9\textsubscript{\textcolor{TealBlue}{$\uparrow$\,0.3}} \\
\checkmark & \checkmark & \checkmark & \textbf{93.0}\textsubscript{\textcolor{TealBlue}{$\uparrow$\,0.8}} & \textbf{91.1}\textsubscript{\textcolor{TealBlue}{$\uparrow$\,1.0}} & \textbf{90.5}\textsubscript{\textcolor{TealBlue}{$\uparrow$\,3.5}} & \textbf{91.8}\textsubscript{\textcolor{TealBlue}{$\uparrow$\,1.9}} \\
\bottomrule
\end{tabular}%
}
\vspace{-10pt}
\label{table:rewardab}
\end{table}

%% file: sec/5_conclusion.tex
\section{Conclusion}
We presented DRS-GUI, a training-free framework that improves GUI grounding by inserting a dynamic region search stage before coordinate prediction. A lightweight UI Perceptor and three perceptual actions guided by an MCTS-based planner and a region quality reward help decouple where to look from what to predict and focus on compact, instruction-relevant regions. Experiments on ScreenSpot, ScreenSpot-V2, and ScreenSpot-Pro show consistent gains over strong general and GUI-specific MLLMs, especially on high-resolution, element-dense interfaces.

\section*{Acknowledgements} This work is supported by the National Natural Science Foundation of China (Grant NO 62376266 and 62406318).